\documentclass[letterpaper, 10 pt, conference]{ieeeconf}  % Comment this line out if you need a4paper
\usepackage{amsfonts}
\usepackage{array,multirow,graphicx}
\usepackage{float}
\usepackage{subcaption}

\IEEEoverridecommandlockouts                              % This command is only needed if 
\usepackage{color}
\usepackage{ulem}
\usepackage{soul}
\usepackage{graphicx}

\title{\LARGE \bf
Lightweight Monocular Depth Estimation via Token-Sharing Transformer
}

\author{Dong-Jae Lee$^{1*}$, Jae Young Lee$^{1*}$, Hyounguk Shon$^{1}$, Eojindl Yi$^{1}$, \\ Yeong-Hun Park$^{2}$, Sung-Sik Cho$^{2}$, and Junmo Kim$^{1}$
\thanks{$^{1}$The authors are with the Department of Electrical Engineering, Korea Advanced Institute of Science and Technology, South Korea. email: \{jhtwosun, mcneato, hyounguk.shon, djwld93, junmo.kim\}@kaist.ac.kr}%
\thanks{$^{2}$The authors are with MOBIS, South Korea. email: \{yhpark0119, sscho\}@mobis.co.kr}%
\thanks{$^{*}$The authors are equally contributed.}
}

\begin{document}
\maketitle
\thispagestyle{empty}
\pagestyle{empty}

%%%%%%%%%%%%%%%%%%%%%%%%%%%%%%%%%%%%%%%%%%%%%%%%%%%%%%%%%%%%%%%%%%%%%%%%%%%%%%%%
\begin{abstract}
Depth estimation is an important task in various robotics systems and applications.
In mobile robotics systems, monocular depth estimation is desirable since a single RGB camera can be deployable at a low cost and compact size. Due to its significant and growing needs, many lightweight monocular depth estimation networks have been proposed for mobile robotics systems. 
While most lightweight monocular depth estimation methods have been developed using convolution neural networks, the Transformer has been gradually utilized in monocular depth estimation recently. 
However, massive parameters and large computational costs in the Transformer disturb the deployment to embedded devices. 
In this paper, we present a Token-Sharing Transformer (TST), an architecture using the Transformer for monocular depth estimation, optimized especially in embedded devices. 
The proposed TST utilizes global token sharing, which enables the model to obtain an accurate depth prediction with high throughput in embedded devices.
Experimental results show that TST outperforms the existing lightweight monocular depth estimation methods. On the NYU Depth v2 dataset,
TST can deliver depth maps up to 63.4 FPS in NVIDIA Jetson nano and 142.6 FPS in NVIDIA Jetson TX2, with lower errors than the existing methods. 
Furthermore, TST achieves real-time depth estimation of high-resolution images on Jetson TX2 with competitive results.
\end{abstract}

%%%%%%%%%%%%%%%%%%%%%%%%%%%%%%%%%%%%%%%%%%%%%%%%%%%%%%%%%%%%%%%%%%%%%%%%%%%%%%%%
\section{INTRODUCTION}
Depth information plays a fundamental and crucial role in various robotics systems and applications such as visual odometry, autonomous driving, robot localization, and visual perception.
Several dedicated sensors such as light detection and ranging, time-of-flight, and structured light are widely used for capturing a depth map.
However, such sensors are costly and bulky, thus making them unsuitable for mobile robotics platforms and edge devices.
In contrast, monocular depth estimation, which estimates the depth map using a single RGB image, can be easily deployed at a low cost and compact size. 
Thus, lightweight deep networks for monocular depth estimation have been widely studied.

Recently, state-of-the-art monocular depth estimation methods \cite{GLPDepth, DepthFormer, DPT} focus on utilizing the Vision Transformer (ViT) \cite{VIT}, which enables the model to learn the global information.
However, the Transformers are usually slower than convolutional neural networks (CNN) due to their massive parameters and quadratic complexity of attention computation. 
This characteristic of the Transformer makes it not applicable directly in real-time tasks. 
Therefore, lightweight deep networks for monocular depth estimation are usually based on CNN \cite{FastDepth, TuMDE, GuidedDepth}, especially for mobile robotics and embedded devices.

To increase the throughput of the Transformer, various methods have been proposed, including redesigning the self-attention \cite{SwinFormer, sparseattention, chen2021psvit} or adopting a CNN-Transformer hybrid architecture \cite{levit, Segformer, Efficientformer, MobileViT, Topformer}.x
The CNN-Transformer hybrid architecture can be categorized into two groups: hierarchy-focused and bottleneck-focused architectures. 
These architectures focused on the resolution of tokens\footnote{For the readers who are not familiar to the Transformer, \textit{tokens} referred to as features, especially for the feature used as the Transformer input.} to reduce the complexity of self-attention. 
The hierarchy-focused architecture gradually decreased the resolution of tokens using the CNN between Transformer, while the bottleneck-focused architecture used only a low-resolution token by placing the Transformer at the end of CNN blocks.
In view of performance, hierarchy-focused architecture shows better accuracy since they can learn the multi-level features containing the global information. However, bottleneck-focused architecture shows better throughput since they apply self-attention only in low-resolution tokens. 

\begin{figure}[!t]
    \includegraphics[width=0.45\textwidth]{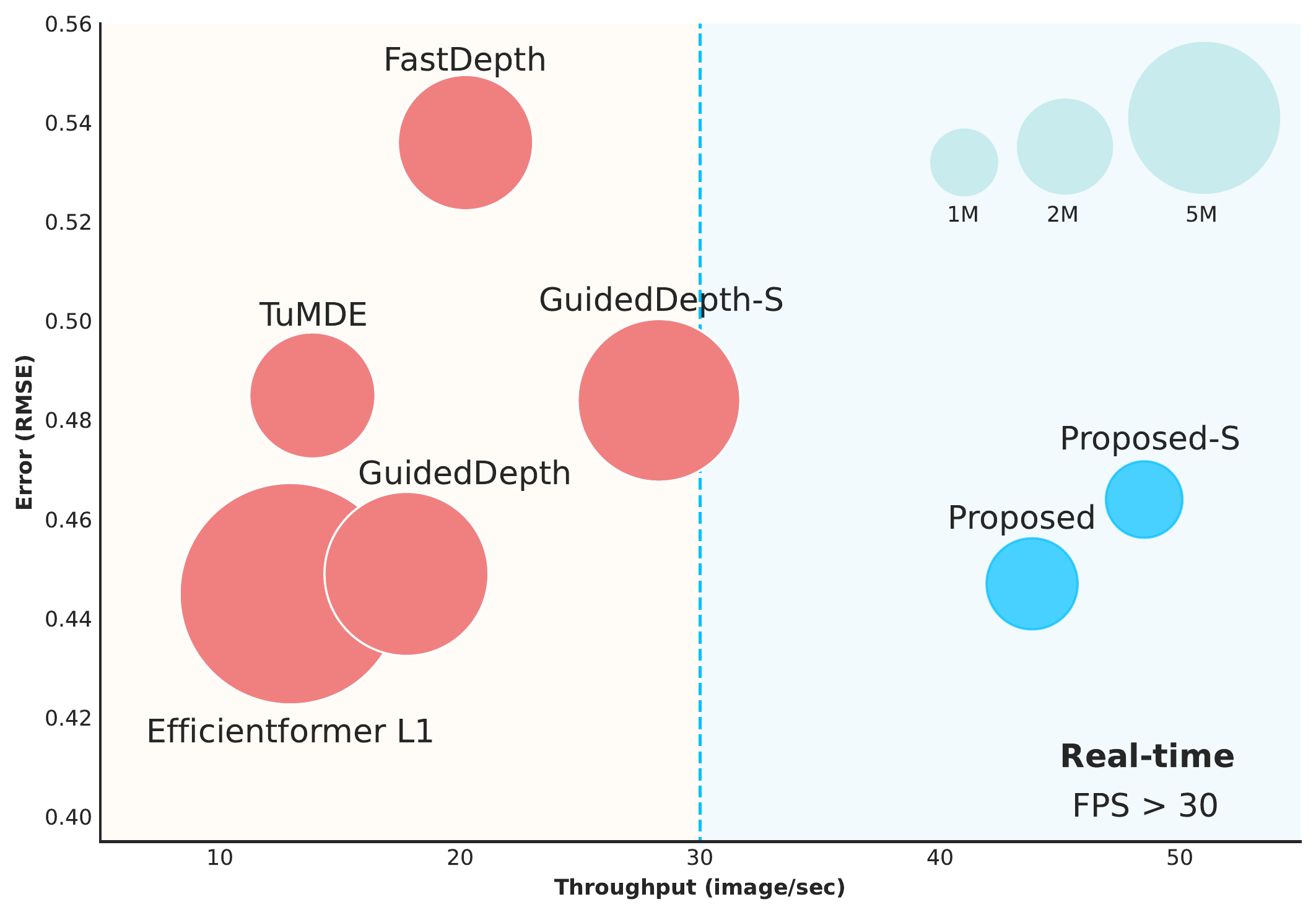}
    \caption{Frame per second (FPS), root mean squared error (RMSE) versus model size on the NYU Depth V2 test set. The existing methods are marked as red and the proposed models (TST) are marked as blue. The throughput is measured with full-resolution $480 \times 640$ input on Jetson TX2.}
\label{fig:intro_preformance}
\end{figure}

\begin{figure*}[!ht]
\centering
    \begin{subfigure}{0.31\textwidth}
        \centering
        \includegraphics[width=\textwidth]{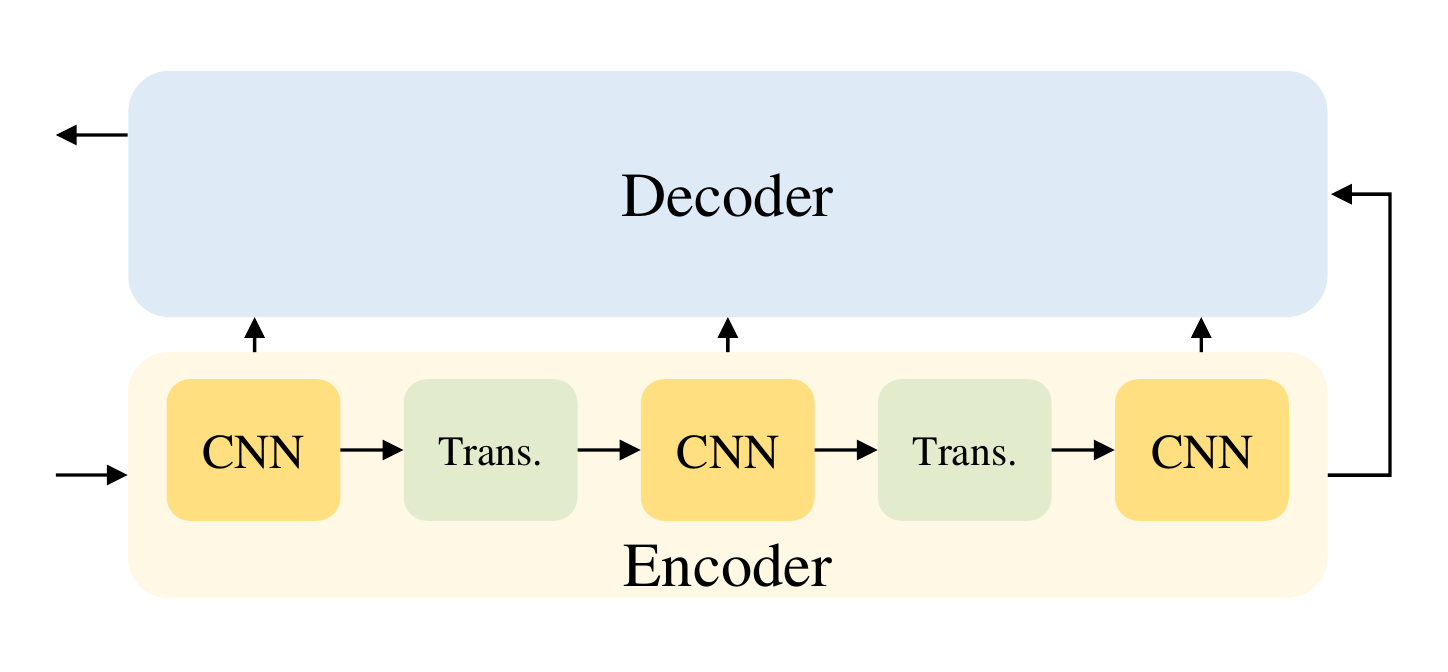}
        \caption{Hierarchy-focused architecture}
        \label{fig:model_archi:alt}
    \end{subfigure}
    \begin{subfigure}{0.31\textwidth}
        \centering
        \includegraphics[width=\textwidth]{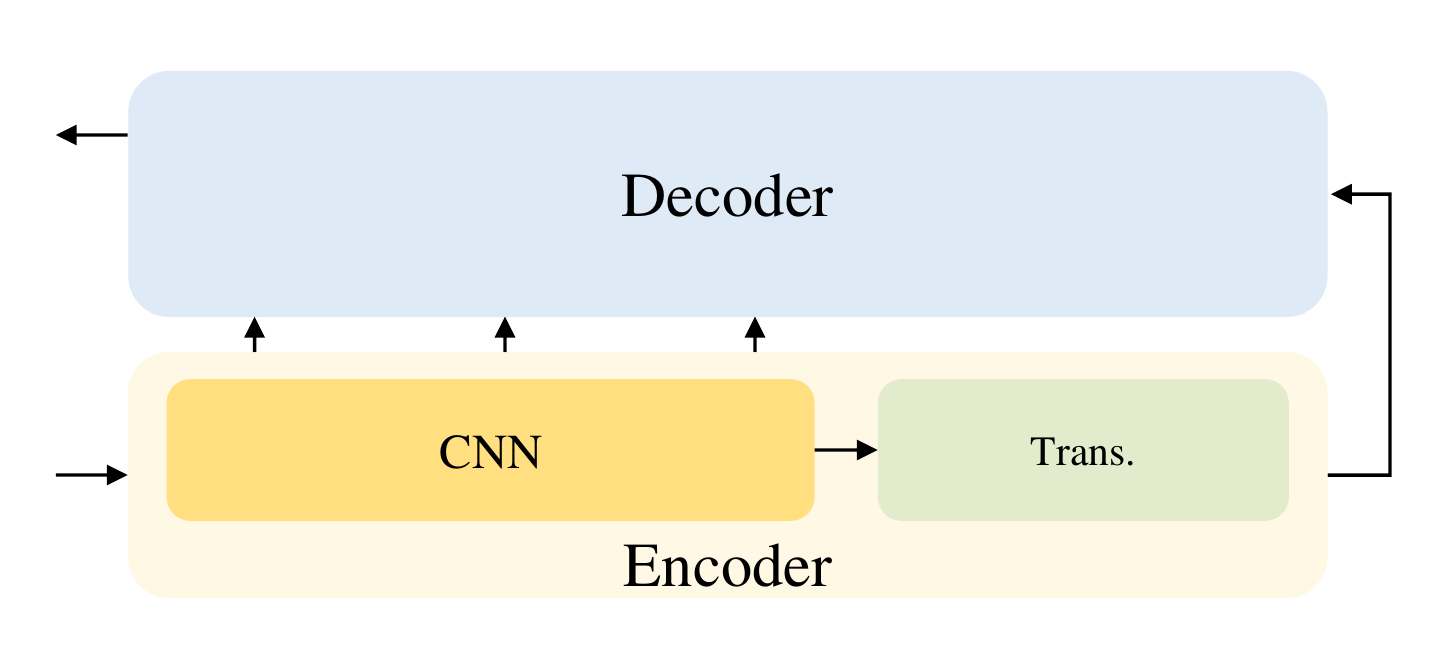}
        \caption{Bottleneck-focused architecture}
        \label{fig:model_archi:cascade}
    \end{subfigure}
    \begin{subfigure}{0.31\textwidth}
        \centering
        \includegraphics[width=\textwidth]{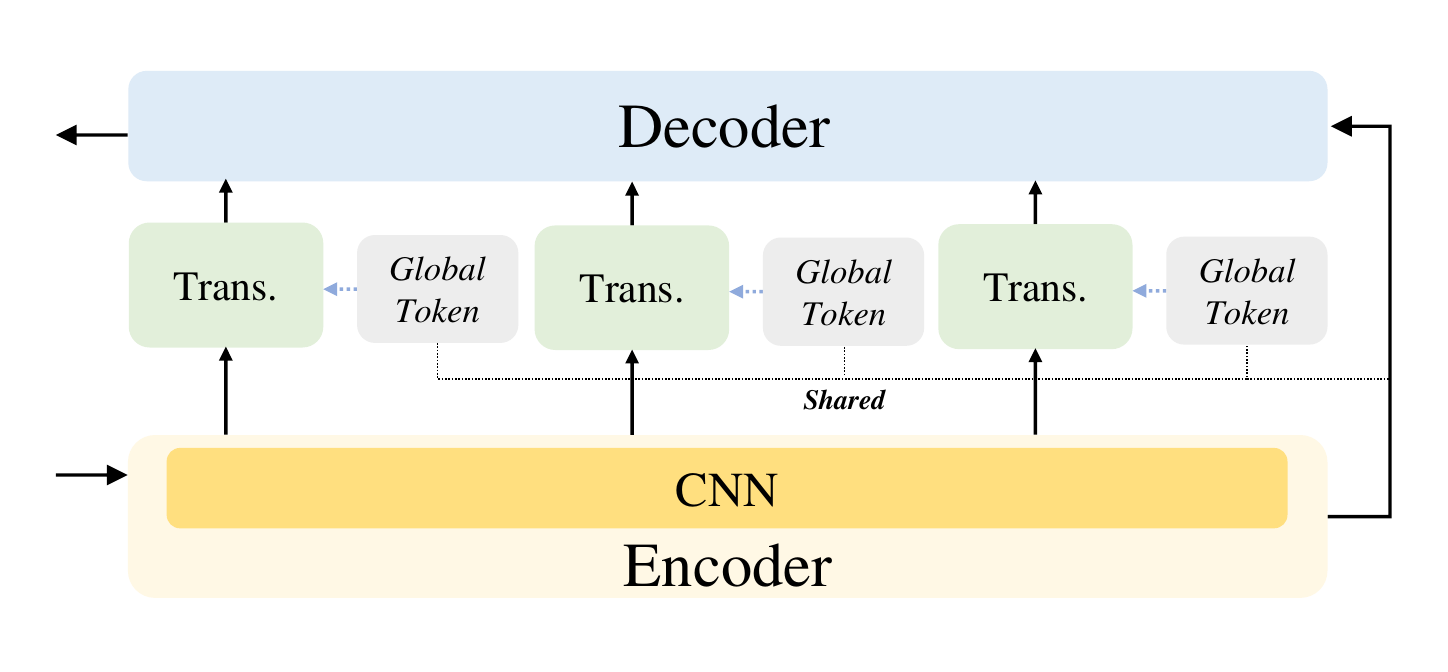}
        \caption{The proposed architecture}
        \label{fig:model_archi:propsoed}
    \end{subfigure}
\caption{Conceptual description of the existing architectures using the transformer and the proposed architecture.}
\label{fig:model_archi}
\end{figure*}

In this paper, we combine the design concepts of hierarchy-focused and bottleneck-focused architectures. 
The main idea in the proposed architecture is \textit{global token sharing}. 
The hierarchy-focused architecture can learn the multi-level features containing global information. However, repetitive self-attention brings significant computational overhead. 
To overcome this limitation, the proposed architecture injects the global information into the multi-level features via the shared global token. Specifically, the global tokens and the multi-level features are extracted from a single lightweight CNN and fed to a cross-attention Transformer. Furthermore, like bottleneck-focused architecture, attention is applied to the low-resolution tokens by downsampling the multi-level features.
Thus, via global token sharing, the model can learn rich global information like the hierarchy-focused architecture, with a high throughput like bottleneck-focused architecture.

To demonstrate the effectiveness of our approach, we experiment with NYU Depth V2 \cite{NYUV2} and KITTI datasets \cite{KITTI}, and further examine embedded systems, i.e., NVIDIA Jetson Nano and Jetson TX2, as well as Titan XP. 
As shown in Fig. \ref{fig:intro_preformance}, the proposed model achieves better results than the existing methods with high throughput. 

\section{RELATED WORK}
\subsection{Vision Transformer}
Based on the success of the Vision Transformer (ViT) \cite{VIT} on the image classification task, diverse architectures using the Transformer have been extensively studied for various applications.
DeiT \cite{DeiT} used distillation learning to reduce the needs of a large dataset.
Swin Transformer \cite{SwinFormer} introduced local-window-based self-attention, which leads to efficient computation of self-attention.
Compact-T \cite{COMPACT-T} and Segformer \cite{Segformer} combined CNN and Transformer to boost their performance and reduce computational costs.
However, it is still incompatible with edge devices because of its high computation complexity.

Recently, several researchers have utilized the Transformer in resource-constrained edge devices.
The existing methods using the Transformer for edge devices can be divided into two categories according to their main focus: 1) reducing the computational cost of the attention mechanism itself \cite{Nystromformer, linformer} and 2) designing lightweight architecture \cite{levit, Segformer, MobileViT, Efficientformer, Topformer}.
To alleviate the computational costs, while the methods in the first category mathematically approximated the attention mechanism, those in the second category proposed hybrid architectures using both CNN and Transformer. 
Our method is grouped into the second category. 
We conceptually summarize the existing lightweight architectures using the Transformer grouped in the second category.

Fig. \ref{fig:model_archi} shows the conceptual description of the existing lightweight architectures and the proposed model.
The existing methods can be subdivided into two groups: 1) hierarchy-focused and 2) bottleneck-focused architecture.
LeViT \cite{levit}, Segformer \cite{Segformer}, and MobileViT \cite{MobileViT} are categorized into the first sub-group (See Fig. \ref{fig:model_archi:alt}).
LeViT simply used the Transformers between convolution layers to reduce the number of tokens in each transformer block.
MobileViT reduced computational costs based on MobileNetV2 \cite{mobilnetv2} backbone with repeated CNN-Transformer blocks. Segformer expanded the usage of CNN layers to overlapped patch merging, which preserves the inductive bias of CNN while fully exploiting the power of the attention mechanism in the Transformer blocks.

Efficientformer \cite{Efficientformer} and Topformer \cite{Topformer} are categorized into the second sub-group (See Fig. \ref{fig:model_archi}). 
Both methods used MobileNetV2 as a backbone for their CNN blocks and utilized an output of CNN blocks as an input of the transformer blocks.
EfficientFormer analyzed various methods using the transformer and proposed several strategies. They observed that the operations of repeated reshaping and permutations in tensors, which are used in the first sub-group, are one of the reasons making the model slow. 
Also, they observed that layer normalization is inferior to batch normalization to optimize the models.
Based on the observations, Efficientformer proposed a model with CNN and transformer blocks that run at MobileNet speed.
Topformer introduced token pyramid pooling, which used the concatenation of feature maps as an input of the transformer blocks, reinforcing the model's representation ability.

To further improve the aforementioned architectures, we propose Token-Sharing Transformer, which is described in Fig. \ref{fig:model_archi:propsoed}. The proposed model achieves the comparable accuracy of hierarchy-focused architecture and throughput of bottleneck-focused architecture. The details of the proposed model and design concept are described in Section \ref{sec:method}.

\subsection{Monocular Depth Estimation} 
Monocular depth estimation aims to predict a depth map from a given RGB image.
Eigen \textit{et al}. \cite{Eigen} pioneered monocular depth estimation using CNN.
Laina \textit{et al}. \cite{LAINA} introduced a fully-convolutional network for monocular depth estimation.
Based on Laina \textit{et al}., various methods have been studied with pre-trained encoders in the classification task as their backbone.
These methods \cite{DPT, DepthFormer, GLPDepth, FastDepth, TuMDE, GuidedDepth} usually focused on designing a decoder network, whereas using the existing networks in the encoder.
% Recently, many studies have focused on the design of encoder architecture using the transformer for depth estimation or that of a decoder to make patch-based architecture adaptable for depth estimation. 
\begin{figure*}[!ht]
\centering
    \includegraphics[width=0.9\textwidth]{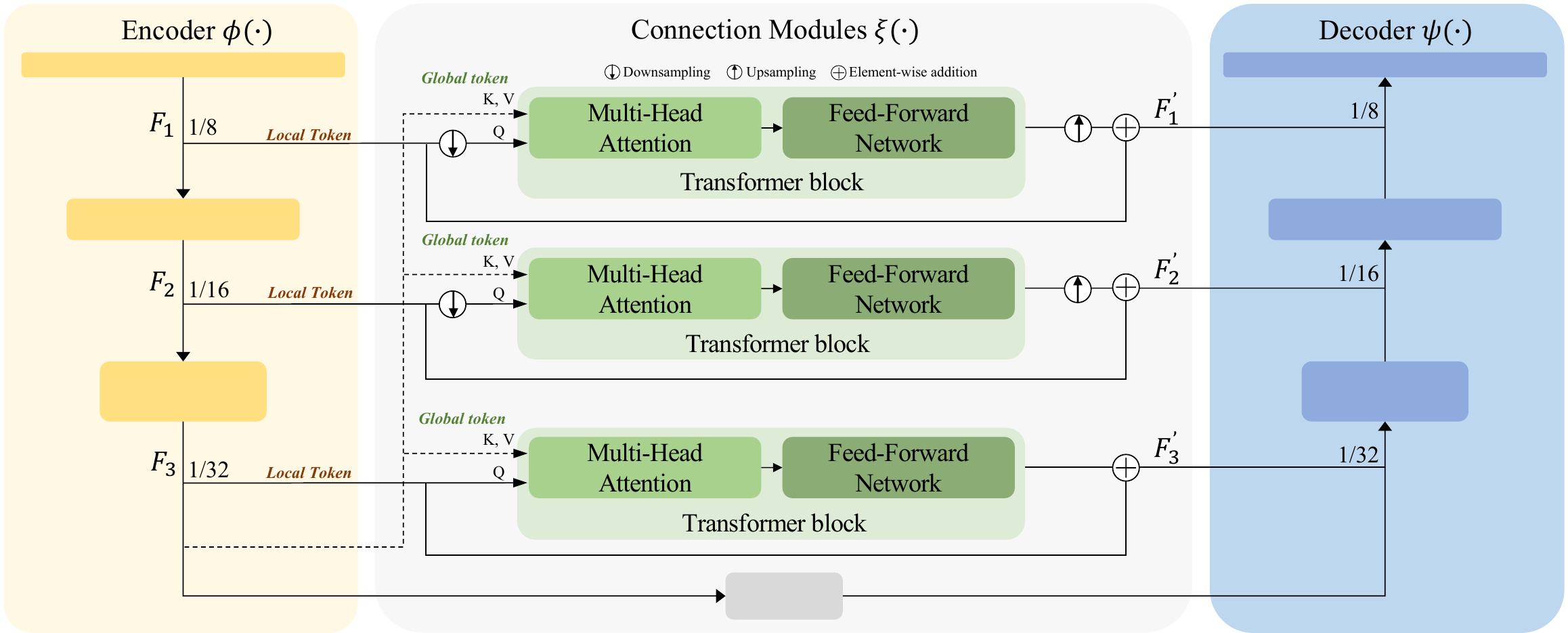}
    \caption{Overall architecture of the proposed model (TST).}
\label{fig:model_prposed}
\end{figure*}
\subsubsection{Transformer-Based Monocular Depth Estimation}
There have been several efforts to use Vision Transformer in Monocular depth estimation. 
DPT \cite{DPT} utilized the global information and large receptive field from ViT encoder.
For their decoder, they use CNN to make a dense prediction. 
DepthFormer \cite{DepthFormer} used two encoders, which consist of transformer and CNN, respectively. 
They introduced the hierarchical aggregation and heterogeneous interaction module to combine the features from the transformer and CNN encoders. 
GLPDepth \cite{GLPDepth} utilized Segformer \cite{Segformer} as their encoder and proposed a selective feature fusion module in their decoder to fuse local and global features.
Since the decoder of GLPDepth is light-weighted and shows reasonable performance, our method utilizes the GLPDepth decoder.

\subsubsection{Lightweight Monocular Depth Estimation}
Since the above networks have heavy complexity for edge devices, several methods have been proposed in lightweight networks for depth estimation.
  FastDepth \cite{FastDepth} utilized MobileNet \cite{mobilenets} as their encoder and proposed a lightweight decoder for depth estimation. 
FastDepth further used network pruning \cite{NetAdapt} to reduce inference time for edge devices. 
Tu \textit{et al}.'s method (TuMDE) used MobileNetV2 as their encoder and a more simplified decoder than FastDepth decoder. 
TuMDE also used a pruning algorithm with reinforcement learning.
Recently, GuidedDepth \cite{GuidedDepth} introduced a guided upsampling block for their lightweight decoder. 
These works utilized compile-time hardware-level optimization such as Tensor Virtual Machine \cite{TVM}, TensorRT \cite{tensorrt}, and quantization on half-precision (FP16) to reduce the computational complexity and inference time.
Following these works, we also use the same hardware-level optimization techniques.

\section{METHOD}
\label{sec:method}

\subsection{Problem Formulation}
Monocular depth estimation aims to learn images to a depth mapping function $f_\theta:\mathbb{R}^ {H \times W \times 3}\rightarrow \mathbb{R}^{H \times W}$, where $\theta$ denotes model weights in $f$. 
Recently, various existing methods have utilized deep neural networks $f$ as a mapping function composed of the encoder and decoder.
To obtain a resulting depth map $\widehat{\mathbf{y}}$ from an input RGB image $\mathbf{x}$, it can be expressed as 
\begin{equation}
    \widehat{\mathbf{y}} = f_\theta(\mathbf{x}) = \psi( \xi (\phi(\mathbf{x}))),
\end{equation}
where $\phi(\cdot)$, $\xi(\cdot)$, and $\psi(\cdot)$ signify the encoder, connection module, and decoder, respectively. 
The encoder $\phi(\cdot)$ extracts the feature maps from the input image and the decoder $\psi(\cdot)$ reconstructs the feature maps into a depth map. 
In general, for $\xi(\cdot)$, skip-connections with simple additions or concatenations have been widely used. In contrast, our Token-Sharing Transformer (TST) is placed in $\xi(\cdot)$, for efficient global information learning.
The details of the proposed architecture are presented in the following sections.

\subsection{Design Concept and Proposed Model}
To design an architecture for monocular depth estimation on edge devices, we start by revisiting the existing lightweight architectures design concept: hierarchy-focused (Fig.\ref{fig:model_archi:alt}) and bottleneck-focused architectures (Fig.\ref{fig:model_archi:cascade}). 

In general, the Transformer based model learns the global information by self-attention. Specifically, query, key, and value are extracted from the token. 
Then, self-attention calculates the similarity between query and key, and recalculates each pixel feature with the weighted sum of the key according to the similarity. 
However, this brings significant computational overhead. 
To reduce the computational complexity, the hierarchy-focused architecture gradually reduces the resolutions of tokens, so that the model can learn the multi-level features containing global information. However, the effect of reducing the complexity is not so significant.
On the other hand, the bottleneck-focused architecture reduces the resolution through CNN and applies self-attention only in low-resolution tokens to remove the computational complexity significantly.
Thus, the hierarchy-focused architecture can achieve high performance, while the bottleneck-focused architecture can achieve high throughput. 
To achieve better performance and throughput simultaneously, our TST combines the two design concepts. 
Specifically, TST focuses on learning the multi-level features containing the global information, with proposed global token sharing.

In the proposed model, we extract the multi-level features from lightweight CNN.
The multi-level features are used as the local tokens, and the high-level feature is used as a shared global token. 
Each Transformer block is composed of Multi-Head Attention (MHA) and Feed-Foward Network (FFN). 
In MHA, TST computes the cross-attention between the shared global token and each local token. 
Since the proposed model utilizes a lightweight CNN for low computational complexity, the low-level and mid-level features have insufficient global information.
However, the cross-attention injects the global information into each local token, thus the model can learn the multi-level features containing the global information. 
Specifically, the query is extracted from the local tokens while the key and value are extracted from the shared global token.
The output features are calculated by the weighted sum of keys according to the similarity between queries and keys. 
To further reduce the computational complexity, the local tokens are downsampled to the size of the global token before cross-attention. 
Then, the output local tokens are fed to FFN with residual mapping for the feature refinement.
After the feature refinement, the outputs of FFN are upsampled to the original token size. 
With residual mapping, the multi-level feature maps containing global information are fed to the decoder to reconstruct the feature maps into the depth map.

\subsection{Overall Architecture}
\label{sec:method:overall}
Fig. \ref{fig:model_prposed} shows the overall architecture of the proposed model in detail.
With the input image $I\in\mathbb{R}^{H \times W \times C}$, where $H$, $W$, and $C$ indicate the height, width, and RGB channels of the image, we extract a set of multi-resolution feature maps $\mathbf{F}=\{ {F}_1, {F}_2, \cdots, {F}_N\}$ through the encoder $\phi(\cdot)$ where ${F}_n\in \mathbb{R}^{\frac{H}{2^{n+2}}\times \frac{W}{2^{n+2}}\times C_n}$.
A set of feature maps $\mathbf{F}$ is fed to the connection layers.
In the connection layers, $\mathbf{F}$ are firstly passed to average pooling layers, so that a set of features maps $\mathbf{F}^{\downarrow}=\{{F}_1^{\downarrow} {F}_2^{\downarrow}, \cdots {F}_N^{\downarrow}\}$, where ${F}_{n}^{\downarrow} \in \mathbb{R}^{\frac{H}{2^{N+2}}\times \frac{W}{2^{N+2}}\times C_n}$, are obtained.
Note that each element in $\mathbf{F}^{\downarrow}$ has the same resolution scaled by $2^{N+2}$.
Each Transformer block in the connection module consists of the multi-head cross-attention (MHA) and feed-forward network (FFN).
In MHA, we utilize each multi-resolution feature map $F_{n}^{\downarrow}$ as queries $Q$ and feature map from the last layer of the encoder $F_{N+1}$ as key $K$ and value $V$.
Then, passing through FFN, the output of the FFN is upsampled to the same resolution with each $F_n$ and added to $F_n$.
The outputs of the connection module $F_{n}^{'}$ are finally fed to the decoder.
The feature map of the last layer in the encoder $F_{N}$ is further fed to the first layer of the decoder passing through a single convolution block.

To customize the proposed model for various devices, we design the base model, coined Token-Sharing Transformer (denoted TST and its small version TST-S). 
For TST and TST-S, the dimension of the multi-level features $\{{F}_1, {F}_2, {F}_3\}$ are set to $\{64, 128, 160\}$ and $\{48, 96, 128\}$, respectively.
For the loss function, we use scale-invariant log loss \cite{Eigen}.

\subsubsection{Token-Sharing Transformer}
In each transformer block, following the observation of Efficientformer \cite{Efficientformer}, we replace the Layer Normalization and GELU activation function \cite{GELU} to Batch Normalization layer and ReLU6 \cite{mobilenets} activation function, respectively.
For MHA, we follow the design as the same as LeViT \cite{levit}, Segformer \cite{Segformer}, and Topformer \cite{Topformer}.
The dimension of queries and keys are set to 16 and that of values is set to 16 and 32, respectively.
The number of the heads $N_{head}$ in each Transformer block is set 2, 4, and 5, which are proportional to the dimension of the input feature map $C_n$ to further reduce the computational cost.
The FFN consists of a single depth-wise convolution with kernel size 3 between two point-wise convolution layers. 
To reduce the computational cost, the dimension of the output feature map from FFN is set to the same as that of the input feature map (i.e., $F_{n}^{\downarrow}$). 

\subsubsection{Encoder and Decoder}
In the encoder, following MobileNetV2 \cite{mobilnetv2} and TopFormer \cite{Topformer}, we use ImageNet \cite{imagenet} pre-trained CNN consisting of Inverted Residual Block.
For details, we follow the setting of TopFormer \cite{Topformer}.
We do not aim to extract rich information or expect a large-receptive field through the encoder.
Instead, we utilize the encoder as a simple feature extractor for tokens.
Thus, the encoder is constructed with shallow layers, which is preferable for the embedded device.
For the decoder, we utilize the decoder of GLPDepth \cite{GLPDepth}. 
Based on the decoder of GLPDepth, we adjust the resolution of feature maps properly for the proposed model.

\begin{figure*}[!ht]
\centering
\begin{subfigure}{0.16\textwidth}
        \centering
        \includegraphics[width=\textwidth]{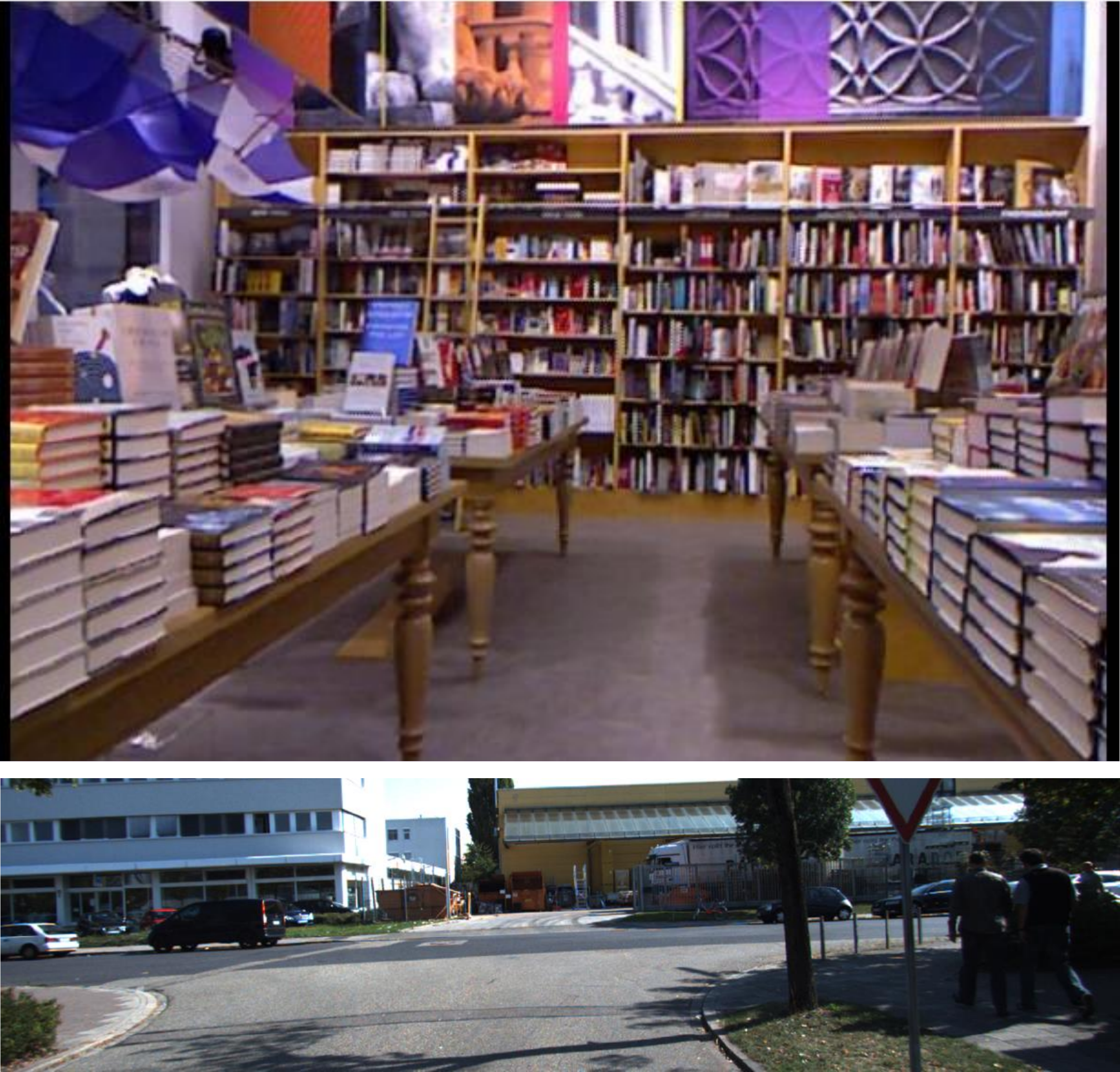}
        \caption{RGB}
        \label{fig:comp_qual_rgb}
    \end{subfigure}
    \begin{subfigure}{0.16\textwidth}
        \centering
        \includegraphics[width=\textwidth]{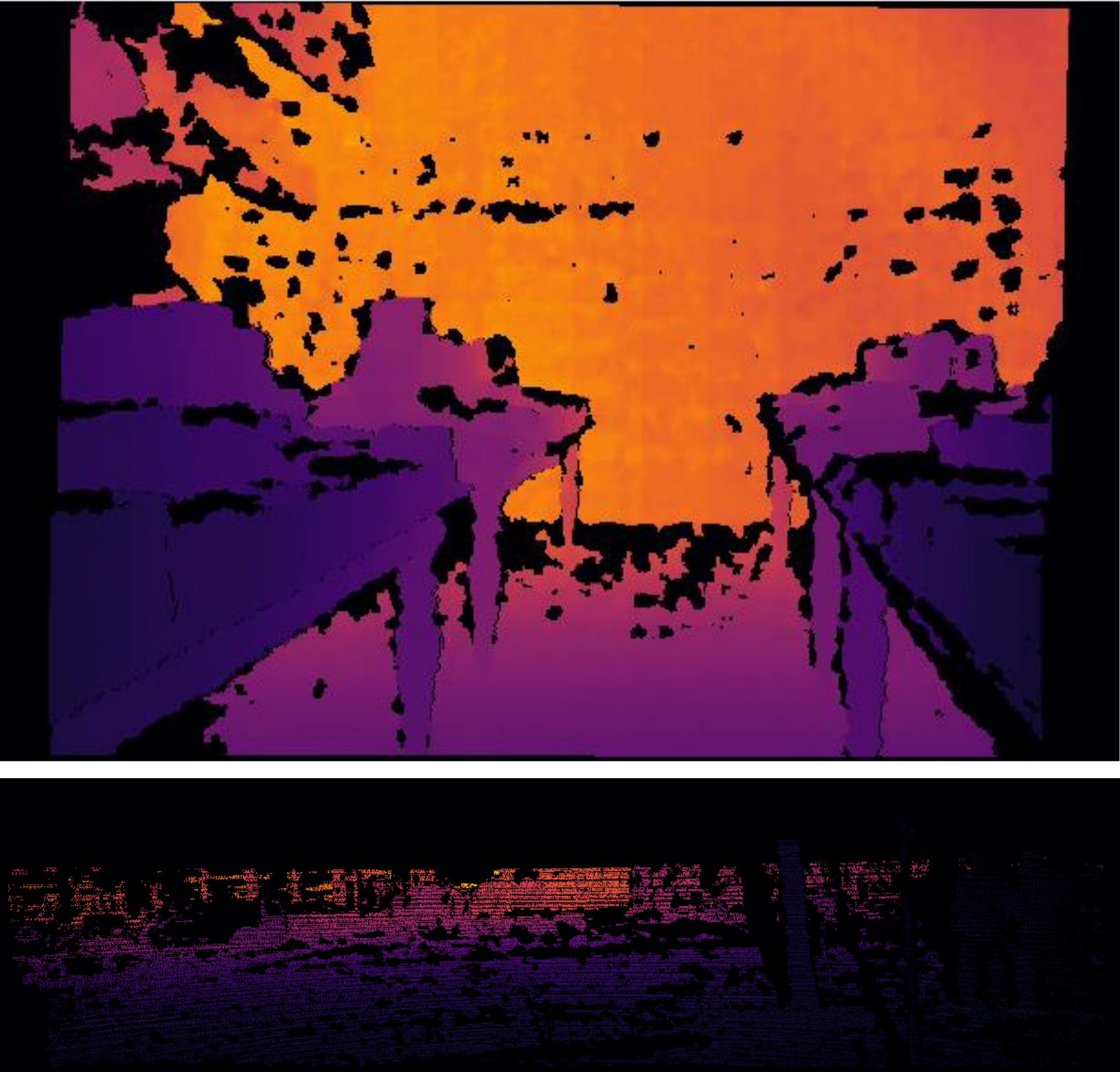}
        \caption{Ground Truth}
        \label{fig:comp_qual_gt}
    \end{subfigure}
    \begin{subfigure}{0.16\textwidth}
        \centering
        \includegraphics[width=\textwidth]{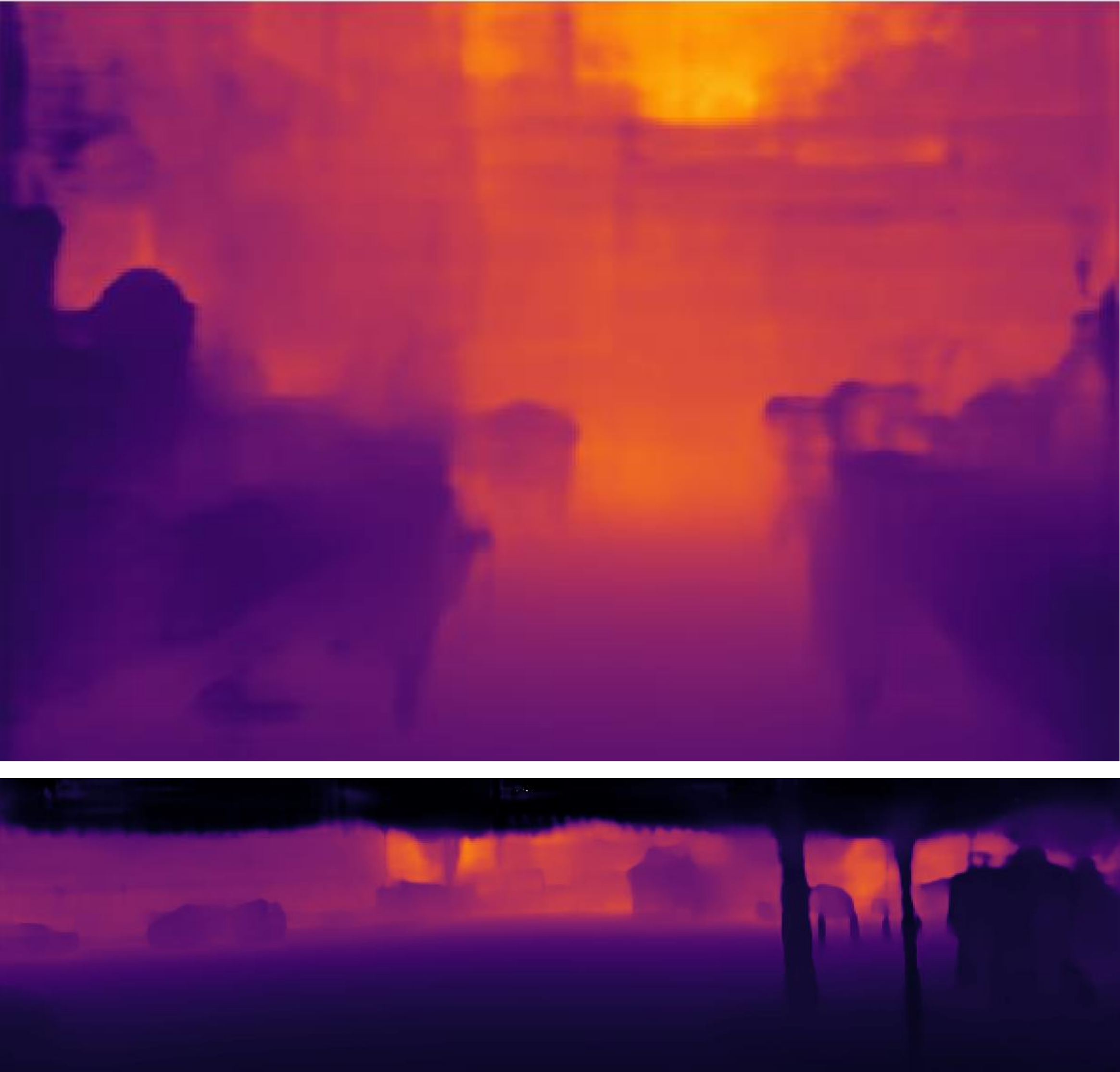}
        \caption{FastDepth \cite{FastDepth}}
        \label{fig:comp_qual_fd}
    \end{subfigure}
    \begin{subfigure}{0.16\textwidth}
        \centering
        \includegraphics[width=\textwidth]{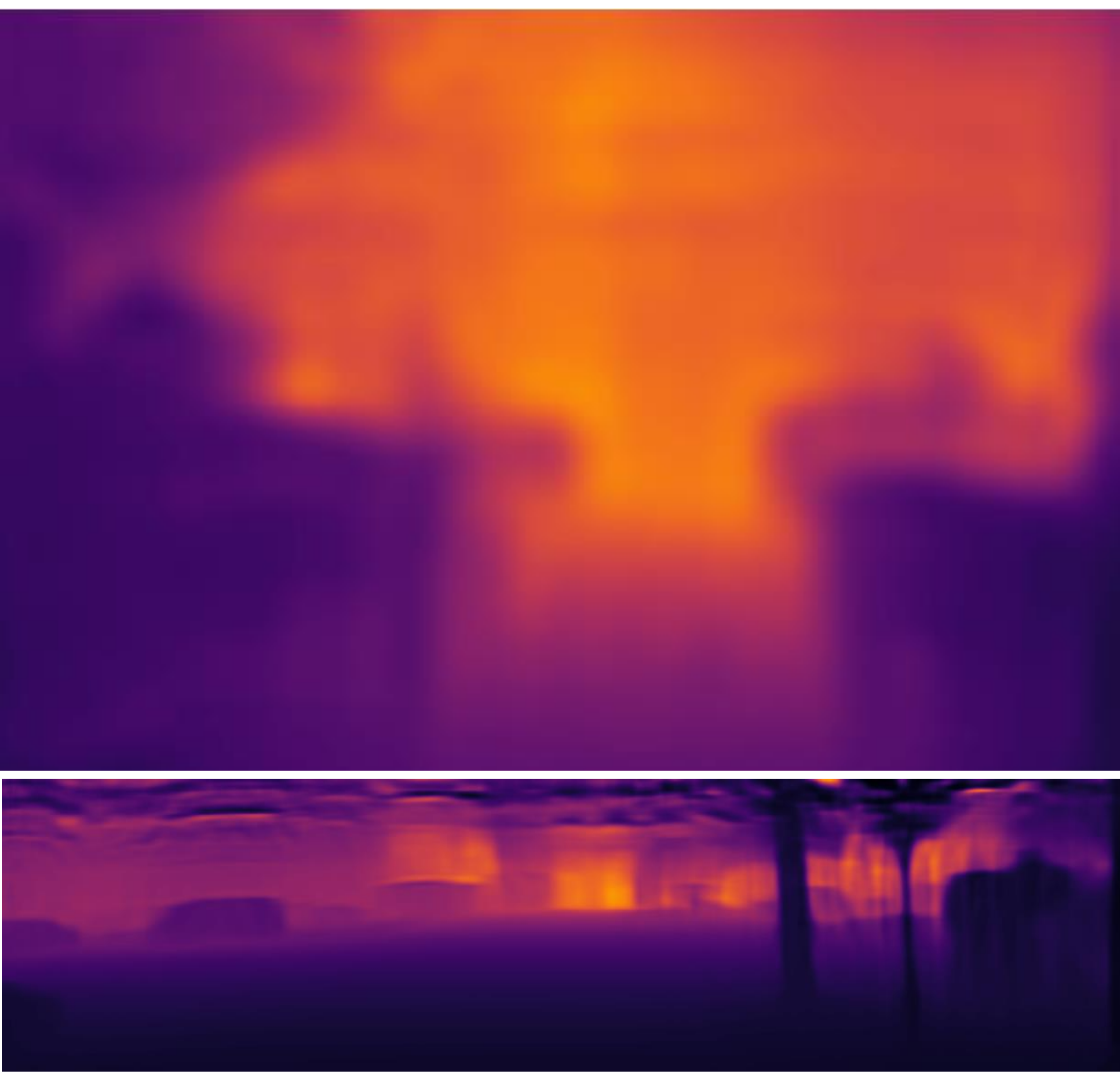}
        \caption{TuMDE \cite{TuMDE}}
        \label{fig:comp_qual_tu}
    \end{subfigure}
    \begin{subfigure}{0.16\textwidth}
        \centering
        \includegraphics[width=\textwidth]{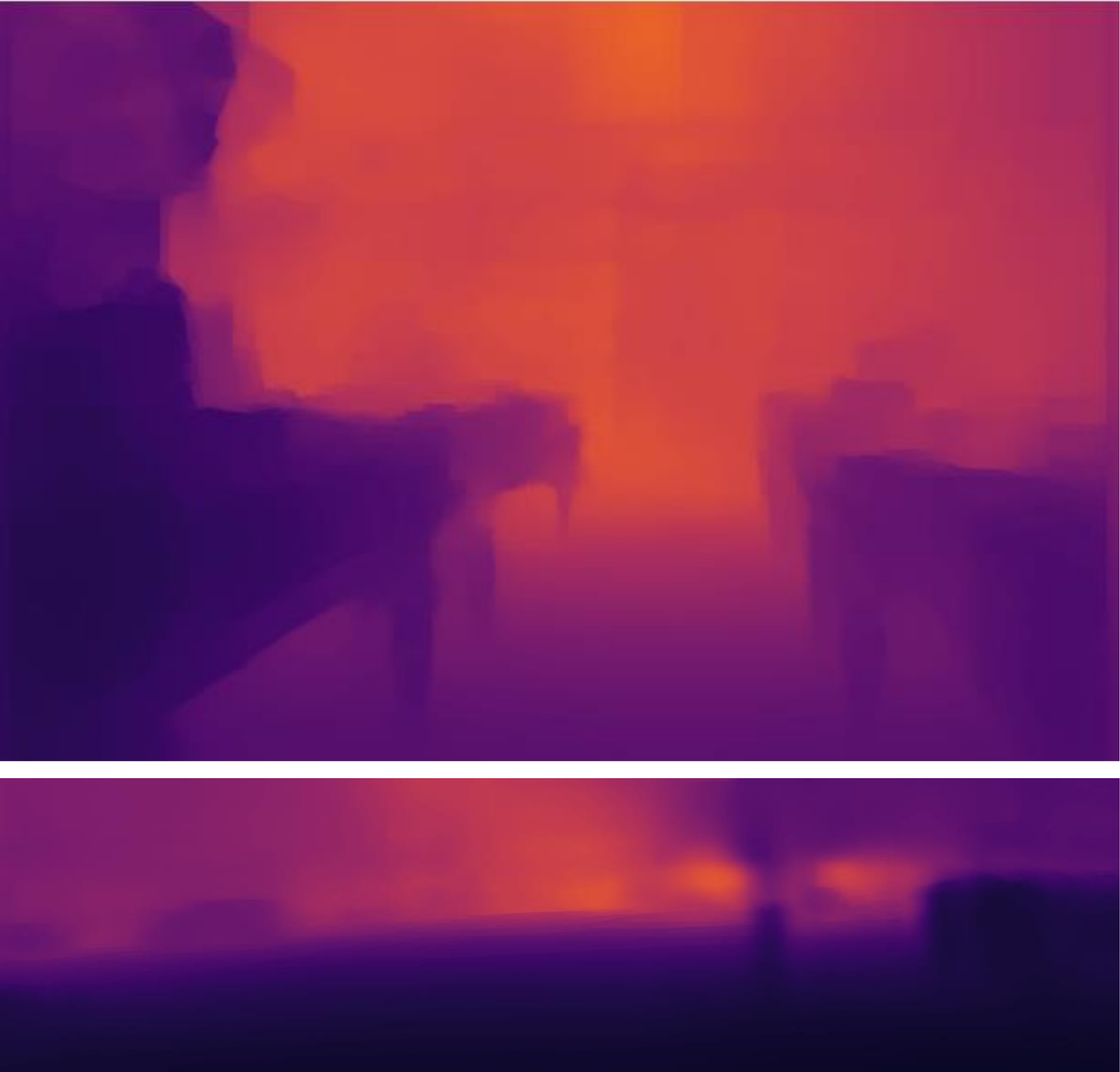}
        \caption{GuidedDepth \cite{GuidedDepth}}
        \label{fig:comp_qual_GD}
    \end{subfigure}
    \begin{subfigure}{0.16\textwidth}
        \centering
        \includegraphics[width=\textwidth]{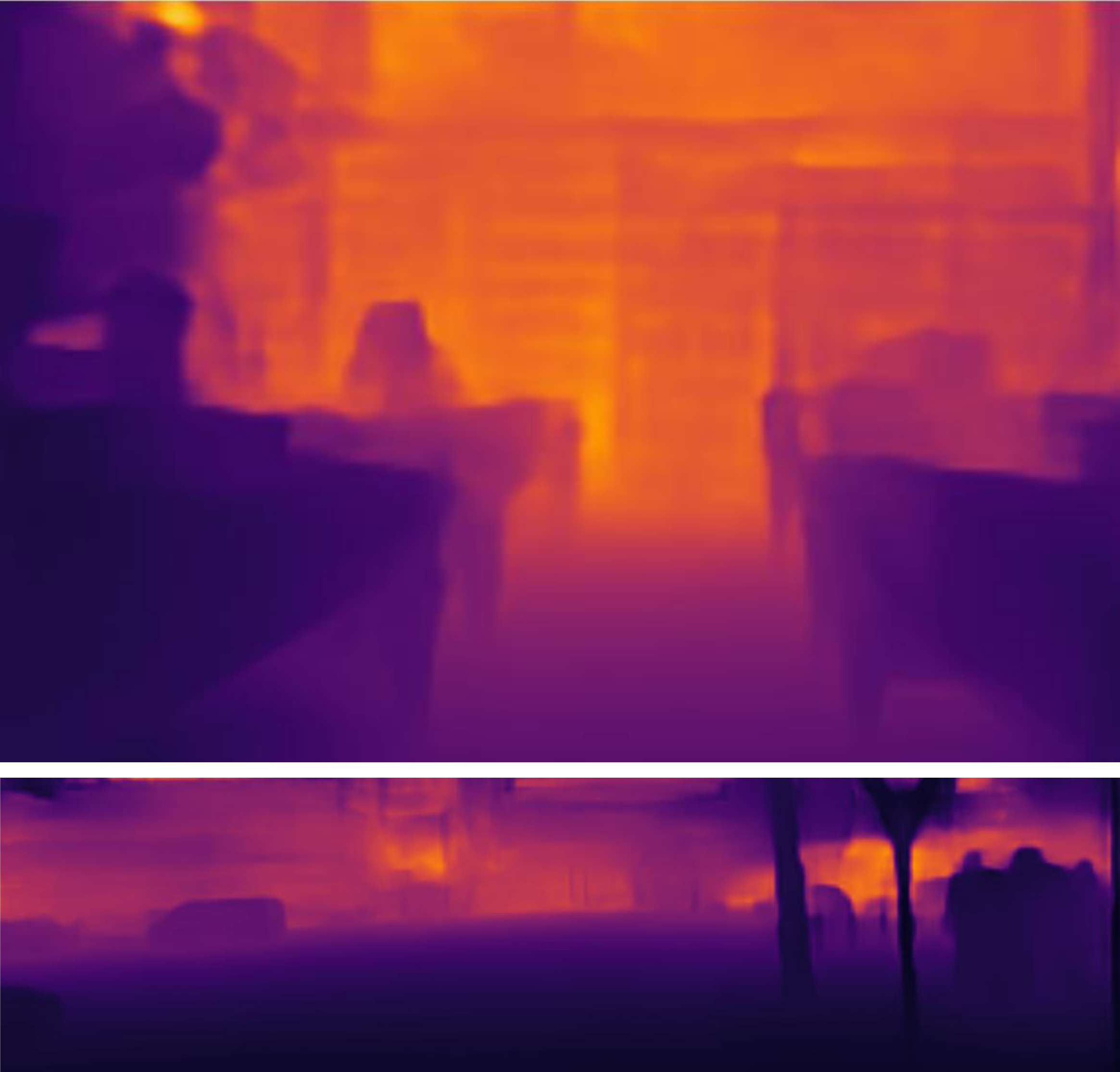}
        \caption{TST}
        \label{fig:comp_qual_pro}
    \end{subfigure}
\caption{Qualitive comparison to the existing methods on NYU Depth V2 ($1^{st}$ row) and KITTI ($2^{nd}$ row) datasets.}
\label{fig:comp_qual}
\end{figure*}
\begin{table*}[t!]
\centering
\caption{Quantitative evaluation on NYU Depth V2 \cite{NYUV2} and KITTI \cite{KITTI} datasets. $\uparrow$ and $\downarrow$ indicate the higher and the lower are better, respectively.}
\label{tab:compa}
\resizebox{0.95\textwidth}{!}{
\begin{tabular}{cclccccccccccccccc}

\hline\hline
\multicolumn{2}{l}{\multirow{2}{*}{}}     			& \multirow{2}{*}{Methods}																				& \#Param.				 				& MACs					 	& \multicolumn{3}{c}{FPS} 											&& \multicolumn{6}{c}{Metrics}     \\ \cline{6-15}
\multicolumn{2}{l}{}                      			&                   																					& (M)									& (G)						& Nano					& TX2				& Titan X 				&& $\delta_1 \uparrow$ 	& $\delta_2 \uparrow$		& $\delta_3 \uparrow$		& AbsRel $\downarrow$	& SqRel $\downarrow$	& RMSE $\downarrow$	\\ \hline\hline
\multirow{12}{*}{\rotatebox[origin=c]{90}{NYU}} 	& \multirow{6}{*}{\rotatebox[origin=c]{90}{$240 \times 320$}}	& TuMDE \cite{TuMDE}   					& 3.44                 					& 1.03                  	& 18.93					& 47.64				& 104.66				&& 0.802 				& 0.953 					& 0.989 					& 0.148 				& 0.106 				& 0.510 			\\
													&                   											& FastDepth \cite{FastDepth}			& 3.96                  				& 1.20                  	& 29.89					& 72.43				& 134.37				&& 0.778 				& 0.943 					& 0.987 					& 0.163 				& 0.121 				& 0.576 			\\
													&                   											& GuidedDepth \cite{GuidedDepth}		& 5.80                 					& 2.63                  	& 24.41					& 55.60				& 70.31					&& 0.823		 		& 0.962 					& 0.992 					& 0.138 				& 0.095					& 0.501 			\\
													&                   											& GuidedDepth-S \cite{GuidedDepth}		& 5.70                 					& 1.52                  	& 38.02					& 91.60				& 71.17					&& 0.787 				& 0.958 					& 0.992 					& 0.146 				& 0.099 				& 0.503 			\\ \cline{3-15}
													&                   											& TST						& 1.80                					& 0.67                  	& 57.15					& 127.85			& 96.33					&& 0.815				& 0.962 					& 0.990 					& 0.143 				& 0.102					& 0.487 			\\
													&                   											& TST-S						& 1.27                 					& 0.56                  	& 63.43					& 142.58			& 96.91					&& 0.802 				& 0.957 					& 0.990 					& 0.148 				& 0.104 				& 0.499				\\ \cline{2-15} 
													& \multirow{6}{*}{\rotatebox[origin=c]{90}{$480 \times 640$}} 	& TuMDE \cite{TuMDE}   					& 3.44                 					& 3,96                  	& 5.05					& 13.84				& 58.40					&& 0.806 				& 0.964 					& 0.993						& 0.143 				& 0.096 				& 0.485 			\\
													&                   											& FastDepth \cite{FastDepth}			& 3.96                  				& 4.69                  	& 8.13					& 20.22				& 87.90					&& 0.781 				& 0.944 					& 0.987 					& 0.157 				& 0.117 				& 0.536 			\\
													&                   											& GuidedDepth \cite{GuidedDepth}		& 5.82                 					& 10.60                  	& 6.51					& 17.76				& 56.67					&& 0.840				& 0.968						& 0.994						& 0.129 				& 0.088					& 0.449 			\\
													&                   											& GuidedDepth-S \cite{GuidedDepth}		& 5.72                 					& 6.10                  	& 10.65					& 28.28				& 62.61					&& 0.817 				& 0.961 					& 0.991 					& 0.140 				& 0.095 				& 0.484 			\\ \cline{3-15}
													&                   											& TST						& 1.80                					& 2.65                  	& 17.03					& 43.83				& 93.21					&& 0.841 				& 0.968						& 0.992 					& 0.132 				& 0.088					& 0.447 			\\
													&                   											& TST-S						& 1.27                 					& 2.21                  	& 18.98					& 48.50				& 94.59					&& 0.828 				& 0.965 					& 0.992 					& 0.136 				& 0.091 				& 0.464 			\\ \hline\hline
\multirow{10}{*}{\rotatebox[origin=c]{90}{KITTI}}	& \multirow{5}{*}{\rotatebox[origin=c]{90}{$192 \times 640$}} 	& TuMDE \cite{TuMDE}   					& 3.44                 					& 1.57                  	& 13.12      			& 33.95      		& 94.08					&& 0.813 				& 0.954 					& 0.987 					& 0.148 				& 0.890 				& 5.282 			\\
													&                   											& FastDepth \cite{FastDepth}			& 3.96                  				& 1.82                  	& 19.39      			& 47.72      		& 116.20				&& 0.808 				& 0.945 					& 0.981 					& 0.150 				& 0.890 				& 5.321 			\\
													&                   											& GuidedDepth \cite{GuidedDepth}		& 5.80                 					& 4.24                  	& 15.97      			& 40.82      		& 67.94					&& 0.857 				& 0.965 					& 0.990 					& 0.119 				& 0.771 				& 4.456 			\\ \cline{3-15}
													&                   											& TST						& 1.80                					& 1.06                  	& 41.25     			& 96.12 			& 94.27					&& 0.866				& 0.974 					& 0.995 					& 0.114 				& 0.649 				& 4.406 			\\
													&                   											& TST-S						& 1.27                 					& 0.89                  	& 46.48      			& 102.30   			& 95.08					&& 0.852				& 0.972 					& 0.994 					& 0.135 				& 0.740 				& 4.621 			\\ \cline{2-15}
													& \multirow{5}{*}{\rotatebox[origin=c]{90}{$384 \times 1280$}}	& TuMDE \cite{TuMDE}   					& 3.44                 					& 6.29                  	& 2.81      			& 7.30      		& 42.55					&& 0.893 				& 0.977 					& 0.995 					& 0.094 				& 0.447 				& 3.705 			\\
													&                   											& FastDepth \cite{FastDepth}			& 3.96                  				& 7.51                  	& 5.03      			& 12.66      		& 55.80					&& 0.889 				& 0.974 					& 0.993 					& 0.094 				& 0.499 				& 3.983 			\\
													&                   											& GuidedDepth \cite{GuidedDepth}		& 5.80                  				& 16.9                  	& 4.12      			& 11.27      		& 46.65					&& 0.868 				& 0.970 					& 0.991 					& 0.115 				& 0.736 				& 4.227 			\\ \cline{3-15}
													&                   											& TST						& 1.80                 					& 4.25                  	& 10.77     			& 26.77 			& 92.34					&& 0.905				& 0.983 					& 0.997 					& 0.087 				& 0.437 				& 3.798				\\ 
													&                   											& TST-S						& 1.27                 					& 3.57                  	& 11.89     			& 29.62 			& 92.57					&& 0.900				& 0.980 					& 0.996 					& 0.089 				& 0.468 				& 3.904 			\\ \hline\hline

\end{tabular}
}
\end{table*}

\section{EXPERIMENTS}
\subsection{Experimental Setup}
% To validate the proposed model, we took several experiments on the NYU Depth V2 \cite{NYUV2} and the KITTI \cite{KITTI} datasets. Furthermore, we perform experiments of different hardware platforms to measure the performance of the proposed network. Additionally, ablation studies on the proposed architecture are conducted to show the effects of the proposed network.
\subsubsection{Datasets} 
\textbf{NYU Depth V2} \cite{NYUV2} consists of $640 \times 480$ indoor images with corresponding depth map captured with Microsft Kinect camera. 
We train the proposed model with approximately 24K, $586 \times 448$ random cropped images and test on 654 images, using the split proposed by Eigen \textit{et al.} \cite{Eigen}.
\textbf{KITTI} \cite{KITTI} consists of approximately 24K images from outdoor driving scenes with spare depth maps captured by the LIDAR sensor. The input images have a resolution of $384 \times 1224$. 
For a fair comparison with GuidedDepth \cite{GuidedDepth}, we resize the RGB image into $384 \times 1280$. 
We use approximately 23K images and 697 images for the test, using the split proposed by Eigen \textit{et al.} \cite{Eigen}.

\subsubsection{Hardware Platforms}
Following the existing methods \cite{FastDepth, TuMDE, GuidedDepth}, we evaluate the model performance on embedded devices: NVIDIA Jetson Nano and NVIDIA Jetson TX2. 
Jetson Nano has a 128-core Maxwell architecture GPU with a Quad-core ARM A57 CPU and 4GB of RAM. 
Jetson TX2 has a 256-core NVIDIA Pascal architecture GPU with Dual-Core NVIDIA Denver CPU, Quad-Core ARM A57 MPCore and 8GB of RAM. 
All evaluation results are reported on 10W and 15W power mode for Jetson Nano and Jetson TX2.

\subsubsection{Implementation Details}
We implement the proposed model on the PyTorch\cite{pytorch} framework. 
For training, we use ADAM optimizer \cite{adam} with customized Cosine annealing warm restarts learning rate scheduler\cite{cosinescheduler}. 
Specifically, for optimizer, we use $\beta_1 = 0.9$, $\beta_2 = 0.99$, \textit{learning rate} $= 0.0003$. 
For scheduler, we use $T_0=10, T_{mult}=2, \gamma = 0.5$. 
We train for 100 epochs and apply the learning rate scheduler for additional 100 epochs. 
For the data augmentation, random horizontal flips, random brightness, contrast, gamma, hue, saturation, and value are used with 0.5 probabilities. Vertical CutDepth  \cite{GLPDepth} is also utilized with 0.25 probability. 

\subsubsection{Evaluation Metrics and Protocol}
\label{sec:exp:evalProtocl}
We use evaluation metrics and protocol, following the existing methods \cite{BinsFormer, DepthFormer, GLPDepth, FastDepth, TuMDE, DPT}. 
For quantitative evaluation, the metrics of $\delta_1$, $\delta_2$, $\delta_3$, absolute relative difference (AbsRel), squared relative difference (SqRel), and root mean squared error (RMSE) are used. 
In the case of resolutions, following resolutions are used for evaluation: full-resolution $480 \times 640 $, $384 \times 1280 $ and half-resolution $240 \times 480$, $192 \times 640$ for NYU Depth V2 and KITTI, respectively. 
Eigen \textit{et al}.'s cropping method \cite{Eigen} is used to evaluate for KITTI dataset.
Note that for evaluation, the predicted depth maps are evaluated after up-sampled to the ground-truth depth map resolution, following GuidedDepth \cite{GuidedDepth}. 
Since GuidedDepth uses customized inpainted dense ground-truth depth maps, we reproduce their results with the evaluation protocol of the other existing methods.
For measurement of throughput in embedded devices, we convert the models to TensorRT engine \cite{tensorrt} with FP16. 
For the measurement in Titan XP, PyTorch with CUDA utility is used without data type conversion. 
For throughput, FPS is measured with 200 samples.

\subsection{Experimental Results}

Table \ref{tab:compa} shows the performance comparisons of the publicly available existing methods and the proposed model (Token-Sharing Transformer, shortly TST) on NYU Depth V2 and KITTI datasets.
Note that TuMDE \cite{TuMDE}, FastDepth\cite{FastDepth}, GuidedDepth\cite{GuidedDepth} are CNN-based methods. Since previous light-weight depth estimation methods are based on CNN architecture, comparsion with Transformer based architecture are presented in ablation study.
``-S'' indicates the small version of each model.
Fig. \ref{fig:comp_qual} shows the qualitative evaluation of the existing method and TST.
\subsubsection{NYU Depth V2}
For the evaluation of the full resolution, TST outperforms the state-of-the-art models for monocular depth estimation in embedded devices in terms of parameter, computation, and inference speed. 
TST achieves better or comparable performance compared to GuidedDepth \cite{GuidedDepth} base model with $3\times$ less parameter (1.9M vs. 5.82M), $4\times$ less computation (2.21G vs. 10.60G), and $3\times$ faster FPS on both embedded devices (17.03 FPS vs. 6.51 FPS, 43.83 FPS vs. 17.76 FPS for Jetson Nano and TX2, respectively) in full-resolution prediction. There are slight performance drops when evaluating the half resolution because we aim to optimize TST to predict the high-quality depth map concerning the input RGB resolution. However, TST achieved comparable performance with much fewer parameters and computation, and faster inference times. 
Furthermore, our small model can achieve 63.43 FPS on Jetson Nano with better accuracy compared to GuidedDepth small in half-resolution prediction. 

\subsubsection{KITTI}
For the evaluation of KITTI, we reproduce the existing methods since they use different resolutions \cite{FastDepth, TuMDE} or inpainted ground truth depth map \cite{GuidedDepth}. 
Note that our evaluation is done without post-processing of the ground-truth depth map. 
Since the model weight of GuidedDepth-S \cite{GuidedDepth} for KITTI is not publicly available, we can not reproduce their results. 
In this experiment, TST outperforms the existing methods in terms of performance, with far more small parameters, computation, and latency.
On Jetson Nano, both TST and TST-S achieve over 40 FPS and over 10 FPS for the half and full resolutions, respectively. 
It is worth noting that TST achieves the highest FPS with the smallest parameter and computation without decreasing the overall accuracy. 

\begin{table}[b!]
\centering
\caption{Ablation studies on the role of the Transformer.}
\label{tab:compa2}
\resizebox{0.45\textwidth}{!}{
\begin{tabular}{lcccccccc}

\hline\hline
\multirow{2}{*}{Methods}	& \multicolumn{4}{c}{Metrics}     															\\ \cline{2-5}
							& $\delta_1 \uparrow$ 	& abs rel $\downarrow$	& sq rel $\downarrow$	& RMSE $\downarrow$	\\ \hline\hline
TST w/o Trans.  			& 0.804					& 0.143					& 0.095					& 0.481				\\
TST w/ self-attn.   		& 0.828					& 0.134					& 0.090					& 0.454				\\ 
TST 						& 0.841 				& 0.132 				& 0.088 				& 0.447 			\\ \hline\hline
\end{tabular}
}
\end{table}

\begin{table}[b!]
\centering
\caption{Experimental results on using the encoders of the existing methods.}
\label{tab:compa1}
\resizebox{0.45\textwidth}{!}{
\begin{tabular}{lcccccccc}

\hline\hline
\multirow{2}{*}{Methods}		& \#Param. 				& MACs 						& \multicolumn{2}{c}{FPS} 					& \multicolumn{1}{c}{Metrics}  \\ \cline{4-6}
								& (M)					& (G)						& Nano					& TX2				& $\delta_1 \uparrow$ 			\\ \hline\hline
Seg. B0 \cite{Segformer}  		& 3.41      			& 6.24                 		& 3.47    				& 8.26      		& 0.845 						\\
Seg. B1 \cite{Segformer}  		& 13.53      			& 24.06                		& 1.67    				& 4.03	      		& 0.863 						\\
Eff. L1 \cite{Efficientformer}	& 10.61     			& 15.88               		& 4.65      			& 12.92      		& 0.842 						\\
Eff. L3 \cite{Efficientformer}	& 30.77     			& 37.65	               		& 2.52	      			& 6.82	      		& 0.872 						\\
Top.-S \cite{Topformer}			& 3.33      			& 2.96              		& 16.53      			& 41.82      		& 0.819 						\\
Top. \cite{Topformer}			& 5.34      			& 3.92                 		& 14.51      			& 36.38      		& 0.836 						\\
TST-S							& 1.27      			& 2.21              		& 18.98      			& 48.50   			& 0.828 						\\ 
TST							& 1.80      			& 2.65             			& 17.03     			& 43.83				& 0.841 						\\ \hline\hline

\end{tabular}
}
\end{table}

\subsection{Ablation Study}
\label{sec:exp:ablation}
In this subsection, we validate our observation and the effectiveness of TST. 
We perform various ablation studies on the NYU Depth V2 dataset. 
All experiments are done in the image resolution $480 \times 640$ and with evaluation protocol described in Section \ref{sec:exp:evalProtocl}.
For a fair comparison, we use the same GLPDepth decoder \cite{GLPDepth} in TST for all experiments.

\subsubsection{Experiments of the role of Transformer}
TST utilizes the Transformer with effective global information injection to local features via cross-attention between multi-level local tokens and global sharing tokens. To investigate the effect of TST, we conduct an ablation experiment with TST without the Transformer and with a self-attention Transformer. 
The results are shown in Table \ref{tab:compa2}. It is obvious that the role of the Transformer is critical to the accuracy of TST. Furthermore, compared to using self-attention, TST with cross-attention can achieve better performance. The results firmly improve the effectiveness of the Transformer. Remaining problem is, how to implement the Transformer for embedded devices. We present the ablation studies on the design concepts of the architectures using the Transformer for the embedded devices in the following section.

\subsubsection{Experiments on different architectures}
Table \ref{tab:compa1} shows the experimental result using different encoders. 
The most light-weighted and second-light-weighted models of each encoder are used for the experiments.
Segformer \cite{Segformer} is categorized into the hierarchy-focused architecture while Efficientformer \cite{Efficientformer} and Topformer \cite{Topformer} are categorized into the bottleneck-focused architecture. In addition, although both Efficientformer and Topformer are categorized as bottleneck-focused architecture, there is a subtle difference between them. 
Efficientformer L1 and L3 focused on increasing overall performance, thus resulting in the model with larger parameters and low throughput compared to Topformer. 
On the other hand, Topformer focused on improving the throughput, thus resulting in sacrificing the accuracy.
In view of the accuracy of the depth prediction, the hierarchy-focused architecture (Segformer B0 and B1) shows better performance than the bottleneck-focused architecture (Topformer). Although Efficientformer L3 achieve higher accuracy compared to Segformer B1, they use $2 \times $ larger parameters. 
In contrast, in view of the throughput, the bottleneck-focused architecture shows better performance than the hierarchy-focused architecture.
The experimental results show that TST successfully combines the two different architectures, i.e., comparable accuracy with $5 \times$ and $4 \times$ throughput in Jetson Nano, compared to Segformer B0 and Effifientformer L1, respectively.

\section{CONCLUSIONS}
In this paper, we propose a lightweight monocular depth estimation method using a Token-Sharing Transformer. 
Token-Sharing Transformer uses multi-level features as local tokens like the hierarchy-focused architecture and uses a single shared token, which has low resolution like the bottleneck-focused architecture, as a global token.
By Token-Sharing Transformer, the proposed model can achieve high throughput without performance drop, which is desirable on embedded devices.
The experimental results show that the proposed model outperforms the existing monocular depth estimation methods in accuracy and throughput. 
The ablation studies show the effectiveness of the Token-Sharing Transformer.
Also, compared to the experimental results using the existing methods as the encoder, the design of the proposed architecture is effective and suitable for the embedded devices.

\bibliographystyle{IEEEtran}
\bibliography{reference}

\end{document}